# Apple Leaf Disease Identification through Region-of-Interest-Aware Deep Convolutional Neural Network

Hee-Jin Yu[0000-0001-7315-7517] and Chang-Hwan Son[0000-0001-7077-3074]

Kunsan National University, South Korea
cson@kunsan.ac.kr

**Abstract.** A new method of recognizing apple leaf diseases through region-of-interest-aware deep convolutional neural network is proposed in this paper. The primary idea is that leaf disease symptoms appear in the leaf area whereas the background region contains no useful information regarding leaf diseases. To realize this idea, two subnetworks are first designed. One is for the division of the input image into three areas: background, leaf area, and spot area indicating the leaf diseases, which is the region of interest (ROI), and the other is for the classification of leaf diseases. The two subnetworks exhibit the architecture types of an encoder–decoder network and VGG network, respectively; subsequently, they are trained separately through transfer learning with a new training set containing class information, according to the types of leaf diseases and the ground truth images where the background, leaf area, and spot area are separated. Next, to connect these subnetworks and subsequently train the connected whole network in an end-to-end manner, the predicted ROI feature map is stacked on the top of the input image through a fusion layer, and subsequently fed into the subnetwork used for the leaf disease identification. The experimental results indicate that correct recognition accuracy can be increased using the predicted ROI feature map. It is also shown that the proposed method obtains better performance than the conventional state-of-the-art methods: transfer-learning-based methods, bilinear model, and multiscale-based deep feature extraction and pooling approach.

**Keywords:** Plant Diseases, Deep Convolutional Neural Networks, ROI.

## 1    Introduction

Information and communications technology has been applied into the existing farming practices to increase the quantity and quality of plants and crops. Smart agriculture sensors including optical sensors, accelerometer, electrochemical sensors, and airflow sensors have been used to measure a leaf's angle and colors, soil properties, pH, soil nutrient levels, etc. [1]. Continuous monitoring yields a vast amount of sensing data, from which a plant diseases and its growth condition can be evaluated through data analysis, thereby enabling an increase in yield while minimizing resources such as water and fertilizer. Particularly, plant disease diagnosis in a timely manner is important to prevent diseases from spreading at an immature state and prevent economic damages to farmers. A large team of experts and farmers can identify plant diseases based on the symptoms on the leaves; however, this manual observation is time consuming and costly. In addition, it is inefficient to continuously monitor all the plants on a large field area. Therefore, the automatic detection of plant diseases is necessary. With the rapid advance in computer vision enabled by deep learning, image-based plant disease detections have garnered particular attention. The deep convolutional neural network (DCNN) [2,3] introduced recently have demonstrated powerful performance for image classification and detection problems. Therefore, image-based approaches have been studied actively using mobile cameras or digital cameras built on autonomous agricultural vehicles for plant disease identification.

## 1.1 Related Works

Regardless of the plant disease, computer vision technologies can be used directly for image-based plant disease identification. To characterize local image appearances, SIFT [4], LBP [5], sparse codes [6], and other handcrafted features [7] including color, entropy, and local homogeneity can be extracted from the preprocessed plant images, and subsequently pooled through bag-of-words (BOW) [8] and Fisher vector encoding (FVE) [9,10] to aggregate those features and obtain image-level representations. Next, given the pooled features, a support vector machine (SVM) [11], which is a data analysis tools, can be trained to classify the plant diseases. Certainly, other tools such as decision trees [12] and dictionary learning [13], can be used for classification. Recently, the DCNN has replaced a series of steps that consist of handcrafted feature designs, pooling, and classification because the DCNN can automatically learn generic representations in a hierarchical manner for discriminative feature extraction. With the emergence of the DCNN, a profound knowledge in feature design, feature pooling, and classification are not necessary, thereby rendering it easier for nonexperts to handle the plant disease identification problems. If a new training dataset is provided, good performance can be obtained through transfer learning, which uses pretrained models such as AlexNet [14], VGG [2], and ResNet [15], and subsequently updates the model's parameters. A large number of studies [16,17,18] have been performed based on transfer learning during the past few years for leaf disease identification. However, the difference is minimal between the transfer-learning-based approaches [16-18] because the architectures used are not new. Therefore, a new region-of-interest-aware DCNN architecture is proposed herein.

It is noteworthy that our goal is to solve the leaf disease identification problem, which is different from the leaf species identification problem [19]. It is necessary to model the leaf shapes for leaf species identification [19, 20]. However, for the leaf disease identification, it is crucial to find the location of the leaf disease to extract discriminative features from background and spot areas separately. Additionally, the test images used for the leaf species identification [19] have solid background colors, whereas in this study, clutter backgrounds including leaves and branches have been considered. Moreover, this paper focuses on the state-of-the-art methods based on deep learning and FVE, and as such, details about traditional approaches are not discussed, for which the reader may refer to related literature [21,22].

### 1.2 Proposed Approach

The primary idea is that leaf disease symptoms can be detected only in the leaf area whereas the background region contains no information about them. During DCNN training, the additional use of a region of interest (ROI) feature map including three areas: leaf area, background, and spot area (i.e., areas with leaf diseases) can provide useful information regarding which features are more important and which features have a decisive role in classifying leaf diseases. Hence, an additional subnetwork that can predict the ROI feature map from an input image is first designed, and subsequently combined with the conventional VGG network to be trained in an end-to-end manner. In other words, two subnetworks exist in the proposed ROI-aware DCNN architecture. One is to predict the ROI feature map that can divide the input images into three areas: background, leaf area, and spot area, and the other is to classify leaf diseases. Compared to the conventional transfer-learning-based methods in [16,17,18], which adopt pre-trained models, in other words, their DCNN architectures exist already, the proposed method suggests a new ROI-aware DCNN architecture. This is a major difference between the proposed method and the conventional transfer-learning-based methods.

### 1.3 Contributions

- In the conventional method [23], a series of steps that consist of image segmentation, feature extraction, and classification are conducted separately. Meanwhile, the proposed method is trainable in an end-to-end manner. In addition, the used segmentation method in [23] can be applied to simple leaf images where background colors are nearly solid. Precisely, the used approach is closer to the color-based clustering algorithm to select predefined colors, thus implying that region boundaries are not provided. In our image database, the background colors are similar to spot colors, i.e., leaf disease colors for a certain disease type; thus, the color-based clustering algorithm might fail to extract the spot colors from the leaf images. In contrast, the proposed method can predict the ROI feature map. Further, this study considers a real environment to some degree in that the leaf images are more complicated in the background than those tested in [23], where the background colors are nearly solid.
- Unlike conventional transfer-learning-based methods [16-18] that already use the existing architectures, the proposed method presents a new ROI-aware DCNN architecture. Particularly, a method of combining a new ROI prediction subnetwork with the VGG subnetwork is introduced. The method to pretrain these two subnetworks separately and subsequently combine them to complete a whole network to be trained in an end-to-end manner is described as well. During the DCNN training, the ROI prediction subnetwork can teach the VGG subnetwork regarding which features in the predicted ROI feature map are more important and which features should have a decisive role for leaf disease identification. In other words, the ROI prediction subnetwork serves as a guide for a more accurate leaf disease identification. Through the experimental results, the effectiveness of the predicted ROI feature map in increasing the recognition accuracy is verified. It is also shown that the proposed method can yield a better performance than state-of-the art methods.

## 2 Conventional Approaches

This section presents an overview of recently introduced state-of-the-art image recognition methods for leaf disease identification [2,16,17,18,23,24,25,26]. This will facilitates readers to understand the differences between the proposed method and the conventional methods.

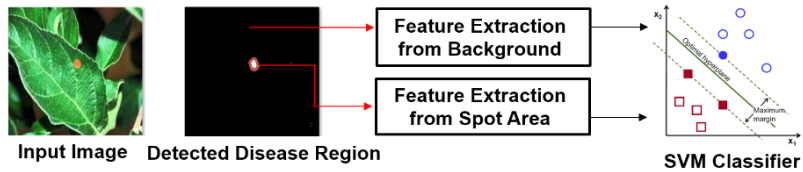

**Fig. 1.** Clustering-based feature extraction for leaf disease identification.

### 2.1 Clustering-Based Feature Extraction

For leaf disease identification, it is important to find the location of leaf diseases. In [23], color-based clustering algorithm is used to define leaf disease colors. More specifically, spot areas with leaf diseases are manually marked and subsequently averaged to define the leaf disease color. Given the test image, leaf disease region is detected by comparing the distance between the predefined leaf disease color and all the pixel colors in the test image. If the distance is smaller than the threshold value, the pixel belongs to the leaf disease region. Subsequently, handcrafted features, such as RGB histogram and LBP [5], are extracted from the background and leaf disease region separately, which are subsequently fed into the SVM classifier. Fig. 1 shows the procedure of the clustering-based feature extraction for leaf disease identification [23]. This approach can work efficiently if the leaf disease colors are significantly different from the leaf colors and background. However, in our image database, leaf colors are similar to spot colors for a certain disease type. This implies that the clustering-based feature extraction might fail to detect the leaf disease region, and this might lead to reduced recognition accuracy.

### 2.2 Deep Feature Extraction and Pooling Methods

Fig. 2 shows the procedure for the multiscale-based deep feature extraction and pooling method (MDFEP) using a pretrained VGG network in [24,25]. Initially, multiscale images are generated to be invariant to scales, and subsequently fed into the pretrained VGG network. Feature maps are extracted from the truncated VGG network at a convolutional layer including nonlinearity (e.g., relu4_3 denoted by [2]). In other words, feature vectors of 512 dimensions can be extracted by gathering all values at the same locations in the feature maps. For example, red circles in Fig. 2 constitute one feature vector. Unlike handcrafted features such as SIFT [4] and LBP [5], the features are learned in a hierarchical manner; thus, these features are referred to as deep features. It has already been verified in [27] that deep features can be generic representations to be applied to different tasks. Next, for all training images in different classes, feature vectors are extracted similarly as mentioned above; subsequently, FVE [9,10] is conducted to pool the feature vectors and characterize the global image appearance based on the Gaussian mixture model (GMM). FVE obtains the gradient vector by taking the derivative of the log likelihood with respect to the GMM's parameters. Instead of FVE, other pooling methods, e.g., BOW [8] and its variant [6] can be used. The FVE feature vectors are fed into the SVM for leaf disease identification. Two choices are available for local feature types and pooling methods, respectively. In other words, either deep features or handcrafted features are used for local feature extraction. Similarly, one of the FVE and BOW can be used for feature pooling. By altering the local feature type in Fig. 2, the discriminative power between the deep features and handcrafted features can be evaluated.

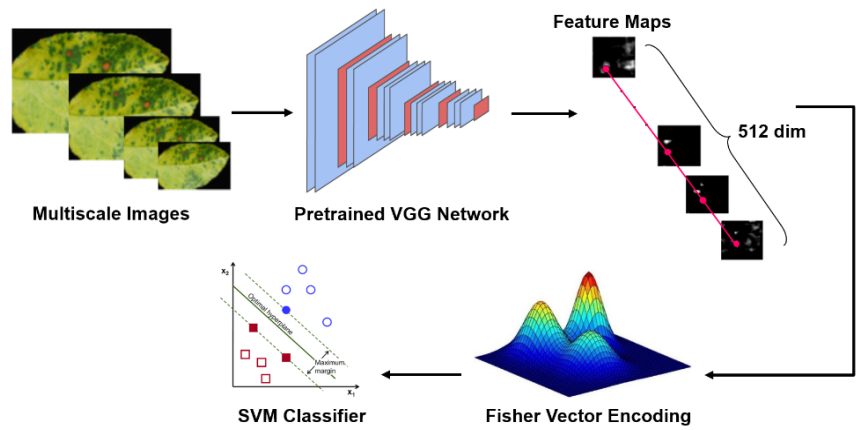

**Fig. 2.** Multiscale-based deep feature extraction and pooling for plant disease identification.

### 2.3 Transfer-Learning-Based Methods

If only the pretrained VGG network, a part of Fig. 2, is applied and subsequently trained with a new training set, the VGG network is finetuned to be adopted to a new task. This type of approach is known as transfer learning (TL) [2]. In this study, a new training set contains leaf disease images and the corresponding label information. Given the new training set, the VGG network can be trained using mini-batch stochastic gradient descent optimization [28]. Instead of the VGG network, ResNet [15] and AlexNet [14] can be used. Compared to the TL method, in the MDFEP method [24,25], as shown in Fig. 2, a series of steps that consist of the local feature extraction, pooling, and classification should be performed. This approach appears more complicated. However, in the TL method, input images should be scaled to a fixed size, thus implying that the TL method can overlook texture information at various scales, i.e., leaf textures and spot areas, which are important for leaf disease identification. Meanwhile, multiscale images are allowed to be used with the MDFEP method. It is worth comparing the performance of the TL method with that of the MDFEP method to demonstrate the effectiveness of using multiscale images.

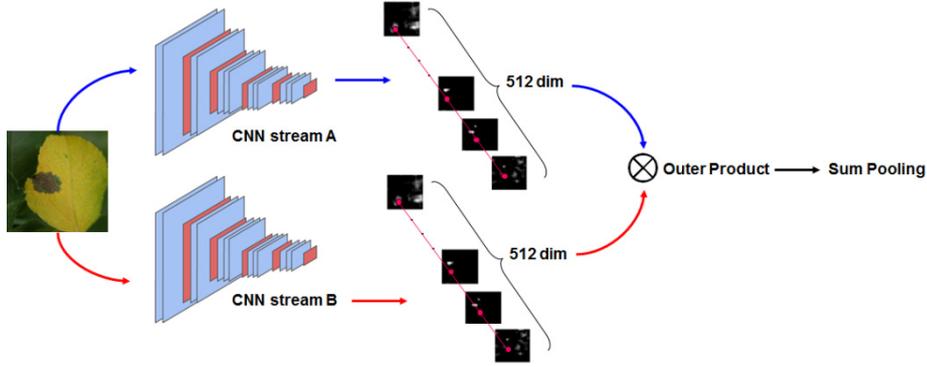

**Fig. 3.** DCNN-based bilinear model for feature extraction and pooling.

### 2.4 DCNN-Based Bilinear Models

Inspired by the success of second-order pooling in visual recognition [29], two-stream CNN architectures are used as feature extractors [26]. Fig. 3 shows the procedure for the DCNN-based bilinear model for feature extraction and pooling. Compared to the MDFEP method using only one DCNN architecture, as shown in Fig. 2, two DCNN architectures are use, as shown in Fig. 3. Similarly, in the MDFEP method, two types of feature vectors can be extracted at the same locations in the feature maps, and subsequently combined using the outer product to model pairwise feature interactions. Next, these combined feature vectors are summed across all locations, thereby producing a global feature vector for an image-level representation. This process is known as sum pooling. In Fig. 3, the two DCNN architectures can be identical or different. In this study, the same VGG network is used to build the two DCNN architectures. For classification, SVM is used, as shown in Fig. 1. The DCNN-based bilinear model has demonstrated powerful performance for fine-grained recognition tasks [26]; thus, it must be tested.

## 3 Proposed ROI-Aware DCNN

Symptoms can be detected only in the leaf area whereas the background region contains no information regarding leaf diseases. Therefore, the additional use of the predicted ROI feature map that contains the leaf area, background, and spot area can teach the DCNN regarding which features in the ROI map are more important and which features should have a decisive role in classifying leaf diseases. Hence, an additional subnetwork to predict the ROI feature map from an input image is designed, and subsequently combined with the conventional VGG network. Fig. 4 shows the proposed

ROI-aware DCNN architecture to identify apple leaf diseases. The proposed architecture consists of two subnetworks: ROI subnetwork and VGG subnetwork. The ROI subnetwork is to transform the input image into the ROI feature map that includes the background, leaf area, and spot area. In the ROI subnetwork, the crop layer applies two-dimensional cropping to the input feature maps. Two input feature maps are required. One is to be cropped and the other is the reference to determine the size of the cropped feature map. The transposed convolution layer applies the transpose of convolution to the input feature maps for upsampling, and $\oplus$ indicates the addition layer that adds the input feature maps by element. The concatenation layer stacks the predicted ROI map on the top of the input image, thereby forming a three-dimensional (3D) input tensor. Through the concatenation layer, the ROI subnetwork is connected to the VGG subnetwork to complete the whole network, as shown in Fig. 4, which is subsequently trained in an end-to-end manner to recognize apple leaf diseases from the 3D input tensor. Other layers such as pooling, rectified linear unit (ReLu), convolution, fully connected, and softmax are described in [2].

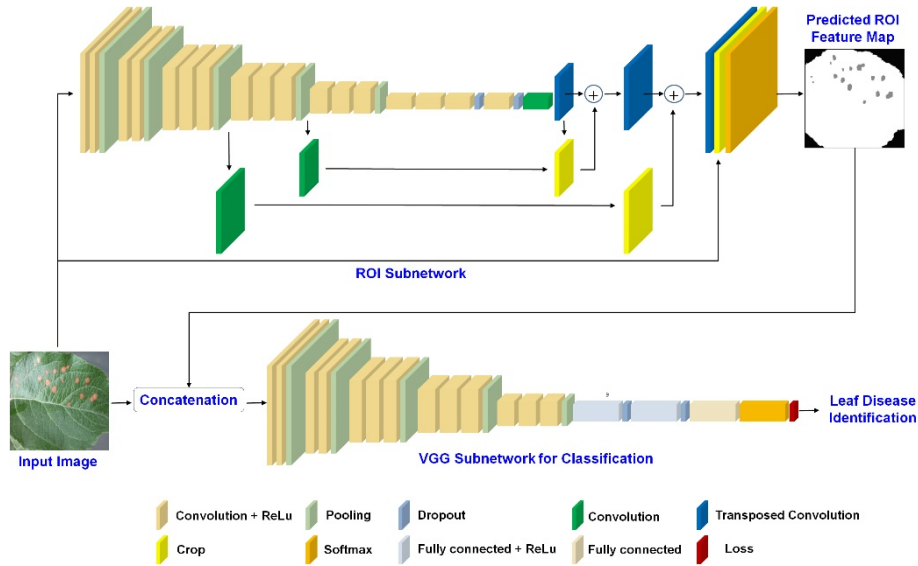

**Fig. 4.** Proposed ROI-aware DCNN for leaf disease identification.

The architecture of the ROI subnetwork in Fig. 4 is inspired by the semantic segmentation in [30]. However, the goal of this study is different from that in [30]; in other words, our goal is not to divide the input image into multiple regions, but to achieve apple leaf disease identification. In the proposed architecture, the ROI subnetwork is pretrained with a new training set that contains ground truth ROI maps; subsequently, this subnetwork is combined with the pretrained VGG subnetwork through the concatenation layer to complete a whole network to be trained in an end-to-end manner. Therefore, the purpose of using the ROI subnetwork is to predict the ROI feature map and subsequently teach the VGG subnetwork regarding which features in the ROI map should have a decisive role in classifying leaf diseases. The ROI subnetwork serves as a guide to achieve a more accurate leaf disease identification. In addition, the ROI feature map is not fixed, in other words, it is iteratively updated during the end-to-end training, which is the main difference between the proposed ROI subnetwork and the semantic segmentation network in [30]. As shown in Fig. 4, the proposed architecture is different from those of TL-based methods [16,17,18] because two types of subnetworks are connected to create a whole network that is subsequently trained in an end-to-end manner. In other words, conventional TL-based methods do not include the ROI subnetwork trainable. If the ROI subnetwork is excluded from Fig. 4, the proposed architecture becomes identical to the conventional VGG network. Therefore, whether recognition accuracy can be increased must be verified by comparing the performance between the proposed ROI-aware DCNN and the conventional VGG network [2]. In Fig. 3, the VGG network is used for leaf disease identification. Certainly, other pre-trained networks such as ResNet and Inception can also be considered.

## 4 Experiments

In this study, the proposed ROI-aware DCNN was implemented using Matlab and trained with four Titan-XP GPUs on a Windows operating system. To compare the proposed method, state-of-the-art methods, i.e., FVE with SIFT [9,10], Clustering-based feature extraction [23], TL methods using VGG and ResNet [16-18], DCNN-based bilinear model [26], and MDFEP [24,25] were tested. Correct recognition accuracy, defined as the ratio of the correctly classified images to the total images, was used for performance evaluation. The number of Gaussians for FVE is 256 and the VGG-16 model in [2] was used as the VGG network in all the methods: TL methods, DCNN-based bilinear model, MDFEP, and the proposed method. The training and testing codes of the proposed ROI-aware DCNN method can be downloaded at https://xxx.xxx.xxx.

### 4.1 Image Collection

All apple leaf images used in this study were provided by the Apple Research Institute in our country. The apple leaf images were categorized into three groups, according to two types of leaf diseases and normal leaf. This implies that a test image was categorized into one of three groups to determine apple leaf diseases. Fig. 5 shows the example of the apple leaf images. The first row shows the normal leaf images, and the second and third rows show the diseased leaf images. Particularly, for the diseased leaf images with marssonia blotch, as shown in the second row, the blotch colors are similar to the normal leaf colors in the background. Therefore, the color-based clustering algorithm [23] might fail to extract the blotch colors from the normal leaves. This reveals that the leaf areas, background, and spot area must be divided. In addition, a real environment was considered to some degree, in that the leaf images were more complicated in the background than those tested in [23], where the background colors were nearly solid. In our database, the total numbers of normal leaf images, diseased leaf images with marssonia blotch, and diseased leaf images with alternaria leaf spot were 118, 120, and 166, respectively.

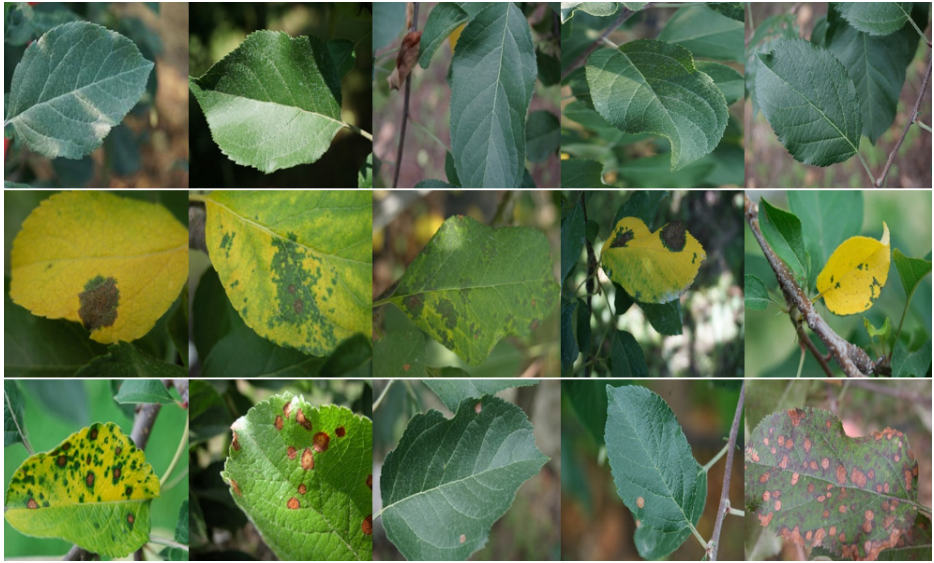

**Fig. 5.** Example of apple leaf images: normal leaf images (first row), diseased leaf images with marssonia blotch (second row), and diseased leaf images with alternaria leaf spots (third row).

### 4.2 Network Training

Before training the whole network in an end-to end manner, as shown in Fig. 4, two subnetworks were first pretrained. To train the ROI subnetwork, ground truth ROI maps are required. In this study, ground truth ROI maps were generated manually through image editing (Photoshop) to divide them into three areas: background, leaf area, and spot area. During image editing, the leaves without diseases are classified by background in the image and those with diseases are classified by leaf area in the image to indicate the ROI. This simplifies the labeling process. Given the ground truth ROI maps, the ROI subnetwork was trained using mini-batch gradient descent optimization [28]. Fig. 6 shows the ground truth ROI maps. The first row shows the leaf images, and the second and third rows show the ground truth ROI maps and the corresponding predicted ROI maps, respectively. Note that the predicted ROI maps are not fixed, i.e., those maps are iteratively updated during the end-to-end training. From these results, it was verified that the ROI subnetwork could provide good performance even though some pixels were misclassified. In our experiment, the mean accuracy of the ROI subnetwork, defined as the ratio of correctly classified pixels to total pixels for each class, was ~86% and mean IoU (Intersection over Union), also known as the Jaccard similarity coefficient, is ~69% [31]. The ratio of training images to testing images was set to 0.5. However, it is noteworthy that the ultimate goal of this study is not ROI prediction but leaf disease identification; this implies that the ROI subnetwork is sufficient to serve as a guide to achieve a more accurate leaf disease identification.

Next, to train the VGG subnetwork, the last three layers of the subnetwork were removed, and subsequently added with a fully connected layer, softmax layer, and log loss layer. Given a new training image set with three labels, the VGG subnetwork was trained using gradient descent optimization. The experimental result indicates that the VGG subnetwork obtains the correct recognition accuracy of 74.7%. Finally, to train the whole network in Fig. 4, the last loss layer of the pretrained ROI subnetwork was removed; subsequently, the softmax layer was connected to the pretrained VGG subnetwork through the concatenation layer that stacked the predicted ROI map on the top of the input image. Subsequently, the whole network was trained with the new training image set in an end-to-end manner. Thus, the ROI subnetwork changed the pretrained parameters to be adopted to a new task. In other words, the predicted ROI map was adjusted during the whole network training for a more accurate leaf disease identification. If the ROI subnetwork were excluded from the whole network, the proposed architecture would be identical to the conventional VGG network. Thus, whether the correct recognition accuracy can be improved by comparing the performance between the proposed ROI-aware DCNN and the pretrained VGG subnetwork must be verified.

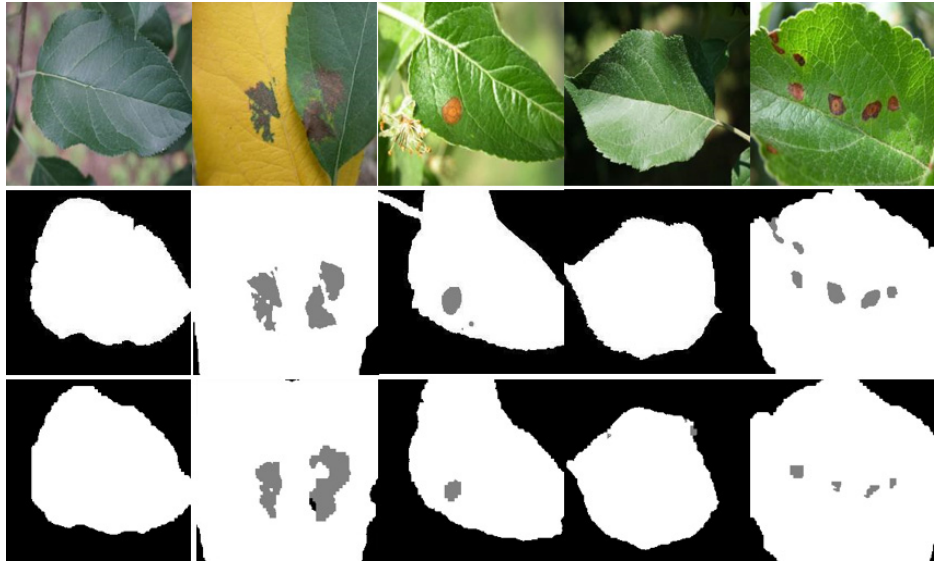

**Fig. 6.** Leaf images (first row), ground truth ROI maps (second row), and predicted ROI maps (third row).

### 4.3 Performance Comparison

Table 1 presents the correct recognition accuracy results for the proposed method and conventional state-of-the-art methods. By comparing the proposed method and the TL method using the VGG network, it is verified that the additional use of the ROI subnetwork increases the recognition accuracy by 9.6%. Hence, it is concluded that the ROI subnetwork can teach the VGG subnetwork regarding which features in the ROI feature map should have a decisive role in classifying apple leaf diseases. In other words, the ROI subnetwork had served as a guide for a more accurate leaf disease identification. It is also shown that the proposed ROI-aware DCNN demonstrates the best performance among all the methods. As shown in Table 1, the performance of the MDFEP method is better than those of the TL methods. This reveals that it is important to use multiscale images for leaf disease identification. In this study, nine scales were used. The TL methods force the size of the input images to be fixed; thus, the leaf texture and spot areas can be removed, thereby resulting in a lower recognition accuracy. The DCNN-based bilinear model demonstrates a better performance than the TL methods using VGG and ResNet networks. This indicates that a pairwise feature interaction model is effective in increasing the discriminative power, thereby improving the correct recognition accuracy. As expected, the performance of the FVE with the hand-crafted SIFT features is worse than that of using deep features. The clustering-based feature extraction method shows the worst performance as the color-based leaf disease detection failed to extract spot colors from the leaf images, owing to the similarity between the both. Additionally, leaf disease detection is sensitive to the threshold value used for clustering. In other words, if the threshold value is high, the background and leaf colors also get included in the detected region. If the threshold value is low, the detected region is too small to extract sufficient features.

**Table 1.** Performance evaluation

| Methods | Correct recognition accuracy |
|---|---|
| Clustering-based feature extraction [23] | 43.9% |
| TL method using VGG network [16,18] | 74.7% |
| TL method using ResNet network [17] | 76.6% |
| MDFEP method [24,25] | 81.7% |
| FVE with SIFT [9,10] | 71.7% |
| DCNN-based bilinear model [26] | 80.6% |
| **Proposed ROI-aware DCNN** | **84.3%** |

## 5    Conclusion

A new ROI-aware DCNN was introduced herein to identify apple leaf diseases. This study was motivated by that leaf diseases existed only in the leaf area and the background contained no information regarding leaf diseases. In the proposed architecture, a new ROI subnetwork to divide input images into the leaf area, background, and spot area was designed and subsequently combined with another VGG subnetwork to be trained in an end-to-end manner. During DCNN training, the ROI subnetwork could teach the VGG subnetwork regarding which features in the ROI map were more important and which features should have a decisive role in classifying leaf diseases. This ROI subnetwork served as a guide for a more accurate leaf disease identification. The experimental results verified that the correct recognition accuracy could be improved using the new ROI subnetwork and by training the whole network in an end-to-end manner. It was also shown that the proposed ROI-aware DCNN yielded a better performance than state-of-the art methods: TL methods, MDFEP method, FVE with SIFT, and DCNN-based bilinear model.

## Acknowledgment

This work was supported by the National Research Foundation of Korea (2017R1D1A3B03030853).